\documentclass[sigconf,pbalance]{acmart}
\AtBeginDocument{%
  }

\acmConference[MM '26]
{Proceedings of the 34th ACM International Conference on Multimedia}
{November 10--14, 2026}
{Rio de Janeiro, Brazil}

\usepackage{multirow}
\usepackage[dvipsnames]{xcolor}
\definecolor{CometRed}{HTML}{FF3B30}
\definecolor{CometOrange}{HTML}{FF8A00}
\definecolor{CometYellow}{HTML}{FFC400}
\definecolor{CometGreen}{HTML}{00C853}
\definecolor{CometBlue}{HTML}{1E88FF}
\newcommand{\ColorfulCOMET}{\textbf{\textit{\textcolor{CometRed}{C}\textcolor{CometOrange}{O}\textcolor{CometYellow}{M}\textcolor{CometGreen}{E}\textcolor{CometBlue}{T}}}}

\setcopyright{none}
\acmDOI{}
\acmISBN{}

\begin{document}

\title[COMET: Motion-Enhanced Temporal Reasoning]{
COMET: Contrastive Motion-Enhanced Temporal Reasoning for
Video Multimodal Large Language Models}

\author{Chenghua Zhu}
\authornote{These authors contributed equally to this work.}
\affiliation{%
  \department{Guangdong Provincial Key Laboratory of Ultra High Definition Immersive Media Technology, Shenzhen Graduate School}
  \institution{Peking University}
  \city{Shenzhen}
  \country{China}
}

\author{Zhaolu Kang}
\authornotemark[1]
\affiliation{%
  \department{School of Software and Microelectronics}
  \institution{Peking University}
  \city{Beijing}
  \country{China}
}

\author{Qifan Shi}
\authornotemark[1]
\affiliation{%
  \department{School of Future Technology}
  \institution{South China University of Technology}
  \city{Guangzhou}
  \country{China}
}

\author{Siyan Wu}
\affiliation{%
  \department{Aberdeen Institute of Data Science and Artificial Intelligence}
  \institution{South China Normal University}
  \city{Foshan}
  \country{China}
}

\author{Kehan Jiang}
\affiliation{%
  \institution{Peking University}
  \city{Beijing}
  \country{China}
}

\author{Lei Wei}
\affiliation{%
  \institution{Peking University}
  \city{Beijing}
  \country{China}
}

\author{Lianyu Hu}
\affiliation{%
  \institution{Nanyang Technological University}
  \city{Singapore}
  \country{Singapore} 
}

\author{Guangyuan Dong}
\affiliation{%
  \institution{National University of Singapore}
  \city{Singapore}
  \country{Singapore}
}

\author{Mingbo Yang}
\affiliation{%
  \institution{Sun Yat-Sen University}
  \city{Guangzhou}
  \country{China}
}

\author{Rui Lu}
\affiliation{%
  \institution{Pingan Technology}
  \city{Shenzhen}
  \country{China}
}

\author{Guibo Luo}
\correspondingauthor
\affiliation{%
  \department{Guangdong Provincial Key Laboratory of Ultra High Definition Immersive Media Technology, Shenzhen Graduate School}
  \institution{Peking University}
  \city{Shenzhen}
  \country{China}
}

\renewcommand{\shortauthors}{Zhu, Kang, Shi et al.}

\begin{abstract}
Video multimodal large language models have advanced significantly, yet fine-grained motion-temporal understanding remains fragile. The core bottleneck is not only sparse frame sampling, but also the lack of a complete temporal modeling pipeline for explicitly representing frame-to-frame change, enabling appearance-motion interaction, and optimizing temporal direction sensitivity. We propose COMET, a temporally grounded framework that systematically strengthens video MLLMs through explicit temporal representation, appearance-motion fusion, and direction-aware optimization. Architecturally, COMET introduces a temporal motion branch built on Taylor frame differences and injects its motion evidence into the appearance stream via temporal attention bias-enhanced cross-attention. For optimization, COMET combines temporal prior distillation with a forward-reverse TC-GRPO stage that turns temporal order into a direct learning signal and strengthens the model's use of directional motion patterns encoded by the temporal motion branch. The method achieves consistent overall improvements with a pronounced motion-temporal bias: on Qwen3-VL-8B, action-centric tasks (STAR, SSv2) improve by 4.9\% on average, temporal reasoning tasks (NExT-QA, CLEVRER, LLaVA-178K) by 2.1\% over BL-GRPO, while static perception tasks (PerceptionTest) remain on par. The same gain pattern also transfers to InternVL2.5-8B, indicating that COMET generalizes across model families.
\end{abstract}

\ccsdesc[500]{Computing methodologies~Computer vision}
\ccsdesc[500]{Computing methodologies~Natural language processing}
\ccsdesc[300]{Computing methodologies~Reinforcement learning}

\keywords{multimodal large language models, video reasoning, temporal understanding, video question answering}

\maketitle

\begin{figure}[t]
  \centering
  \includegraphics[width=\columnwidth]{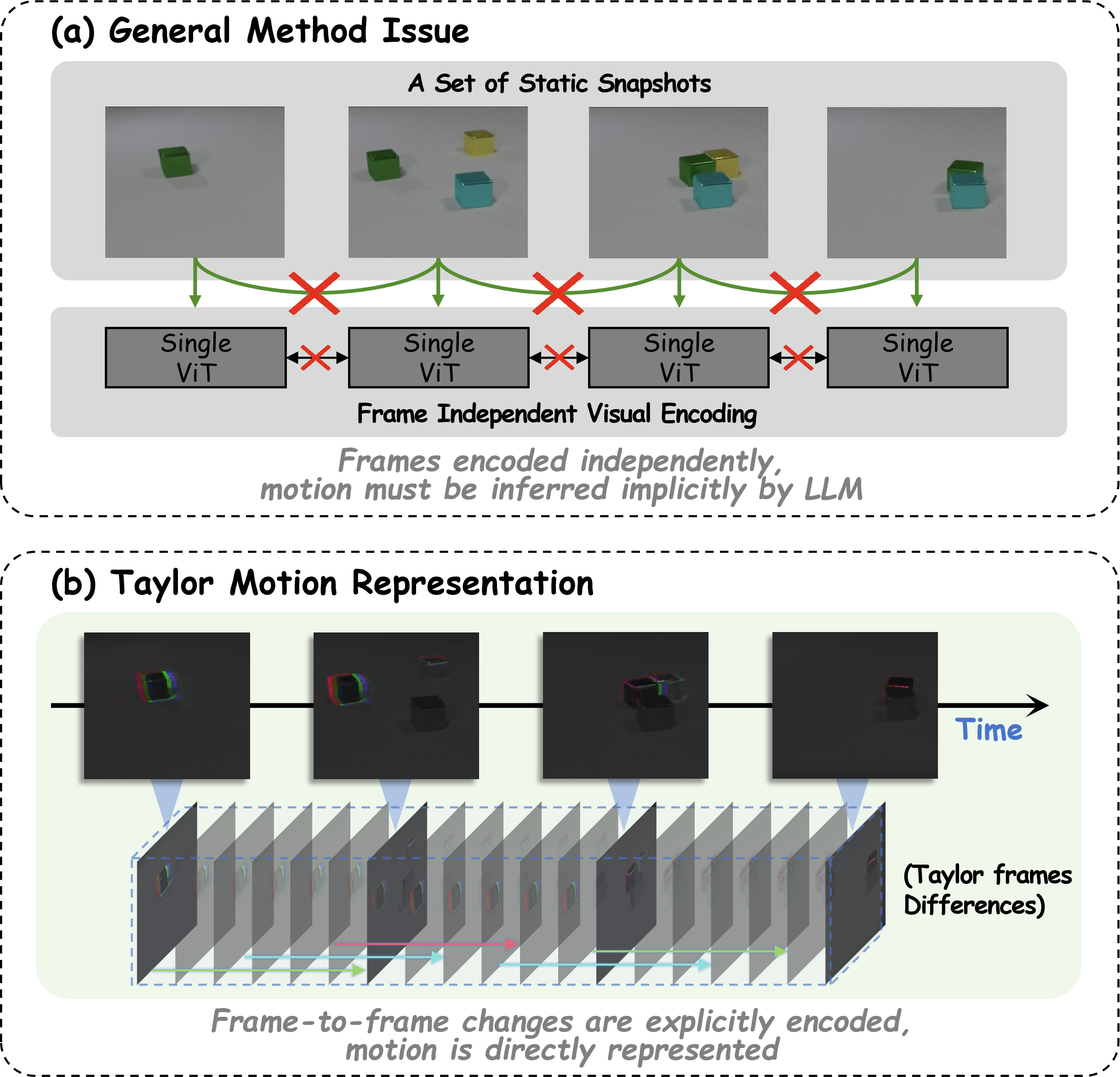}
  \caption{Frame-independent encoding versus explicit temporal representation. (a) sparse RGB frames leave motion implicit. (b) frame-to-frame change makes temporal evidence explicit.}
  \Description{A comparison between frame-independent video encoding and an explicit representation of frame-to-frame temporal changes.}
  \label{fig:limitation}
\end{figure}

\section{Introduction}
\label{sec:introduction}

Video multimodal large language models (video MLLMs), represented by systems such as Qwen3-VL~\cite{qwen3vl2025}, InternVL2.5~\cite{internvl25_2024}, and VideoLLaMA~2~\cite{videollama2_2024}, have recently shown strong progress on video question answering, dense video understanding, and long-form reasoning. Yet their temporal reasoning remains fragile. Many systems answer correctly when static appearance cues or language priors are sufficient, but they still struggle when the answer depends on how a scene evolves over time, such as motion direction, action order, or changes in state. In such cases, temporally grounded evidence is often weaker than the final answer confidence.

We argue that a major reason is that current video MLLMs still underuse video as a temporal signal. In many pipelines, videos are reduced to sparsely sampled static frames~\cite{tsn2016} that are then encoded independently, often as sparse frame tokens~\cite{timesformer2021}, before being compressed, pooled, or aligned with the language model. Explicit inter-frame interaction is absent, so temporal evidence becomes sparse and discontinuous. As a result, video data utilization remains low, temporal coherence is poorly preserved, and downstream reasoning can fall back to appearance-dominated shortcuts.

This limitation is illustrated in Fig.~\hyperref[fig:limitation]{1 (a)}: frame-independent encoding leaves motion to be inferred implicitly, while explicit frame-to-frame change representation exposes temporal evidence directly.

More broadly, the problem lies not in a single missing module, but in a mismatch at two levels: architecture and optimization. Architecturally, frame-to-frame change is not represented explicitly, and appearance and motion do not interact effectively across aligned frames, making temporal cues difficult to incorporate into visual reasoning. At the optimization level, training does not directly reward direction- or order-sensitive behavior, so the model can still ignore temporal cues even when relevant signals are present. Improving video MLLMs therefore requires both a temporally grounded architecture and an optimization strategy matched to temporal sensitivity.

Based on this view, we propose \textbf{COMET} (Contrastive Motion-Enhanced Temporal Reasoning), a temporally grounded video MLLM framework that addresses both levels. Architecturally, we instantiate the temporal branch with a lightweight Taylor-style motion signal that preserves motion directionality (as illustrated in Fig.~\hyperref[fig:limitation]{1 (b)}), and fuse it with the appearance branch via temporal attention bias (TAB)-enhanced cross-attention, so that explicit motion evidence becomes available before entering the language model. For optimization, we first distill temporal reasoning traces through supervised learning (TPD), and then introduce a temporal-contrast GRPO stage (TC-GRPO) in which forward and reversed videos create a structured directional contrast for training and encourage the model to use directional motion patterns more faithfully. Rather than relying on a heavy auxiliary motion estimator, COMET makes temporal evidence explicit throughout the pipeline: the motion branch exposes frame-to-frame change, inter-frame fusion injects it into appearance reasoning, and temporal-contrast reinforcement learning turns direction sensitivity into a directly optimized objective.

Our main contributions are fourfold. First, we propose COMET, a unified framework for temporally grounded video reasoning that systematically improves how video MLLMs represent, interact with, and optimize temporal evidence. Second, we design a dual-branch visual encoder that disentangles appearance and motion, using an RGB branch for spatial content and a temporal motion branch based on Taylor frame differences for explicit frame-to-frame change modeling. Third, we introduce TAB-enhanced cross-attention fusion, which guides appearance features to retrieve motion-salient temporal evidence through explicit inter-frame interaction. Fourth, we develop a direction-aware training strategy based on forward-reverse video contrast, turning temporal order into a direct learning signal and strengthening the model's use of directional motion patterns encoded by the temporal motion branch.

\begin{figure*}[t]
  \centering
  \includegraphics[width=0.88\textwidth]{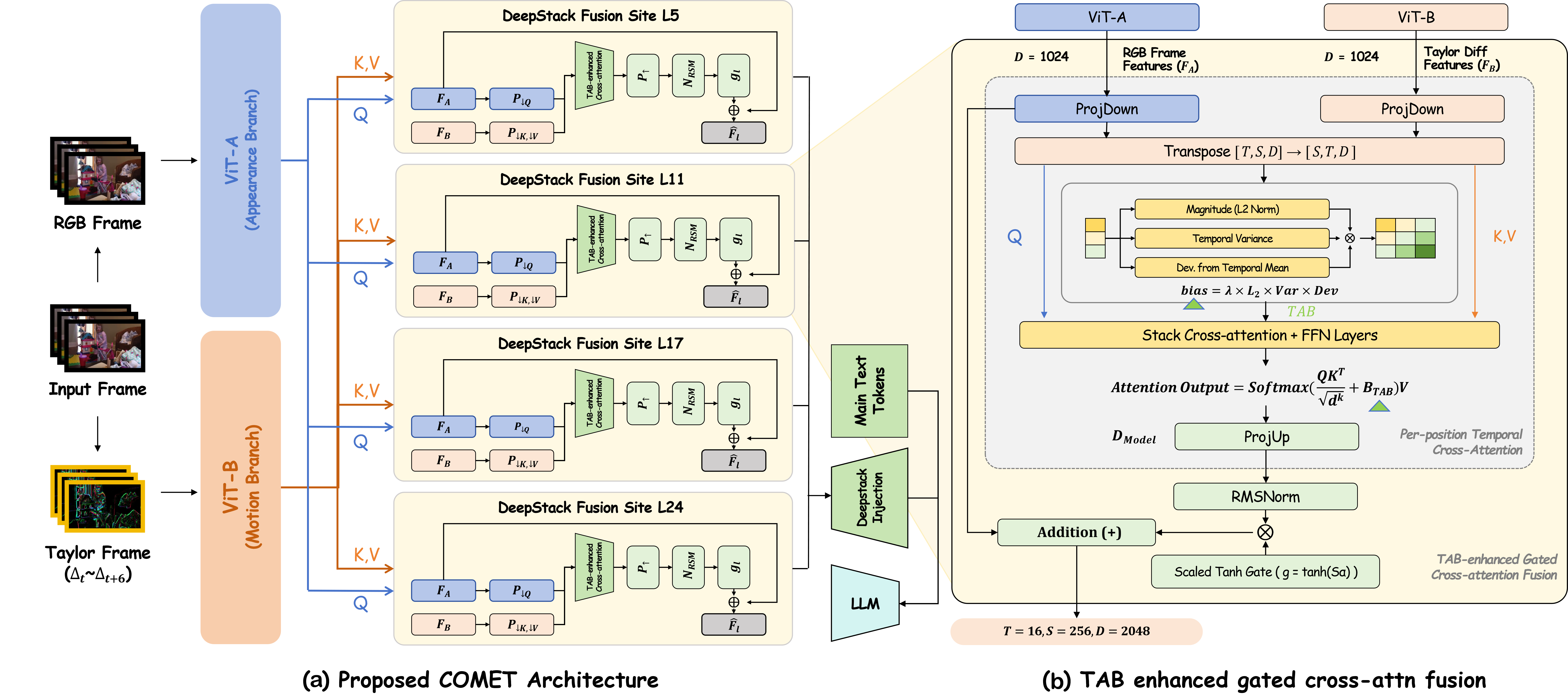}
  \caption{Overview of COMET instantiated on Qwen3-VL-2B. (a) dual-branch architecture with DeepStack fusion. (b) TAB-enhanced gated cross-attention.}
  \Description{An overview of the COMET framework, including its dual-branch visual encoder, multi-level fusion, and motion-aware cross-attention module.}
  \label{fig:overview}
\end{figure*}

\section{Related Work}
\label{sec:related}

Recent video MLLMs typically extend image-pretrained vision encoders and large language models to the temporal domain through sparse frame sampling~\cite{tsn2016}, spatiotemporal token encoding~\cite{timesformer2021}, lightweight temporal adapters~\cite{tsm2019}, or spatiotemporal connectors~\cite{videollama2_2024}. More recent efforts further strengthen temporal perception from the vision side, for example by inserting stacked temporal attention into the encoder~\cite{sta2025} or introducing dedicated temporal fusion modules such as TempFlex's Temporal Fiber Fusion~\cite{tempflex2025}. This line of work has established strong baselines for video question answering and video reasoning, but these methods still mainly improve temporal sensitivity within the existing sparse-RGB pipeline. A more systematic temporal modeling pipeline spanning explicit representation, inter-frame interaction, and training objective design remains underexplored.

Another relevant line of work introduces explicit motion representations into video models. In traditional computer vision, optical flow~\cite{twostream2014,raft2020} and frame differencing have long been used to complement RGB features when dynamic evidence is difficult to recover reliably from static frames alone. In the video MLLM setting, Flow4Agent~\cite{flow4agent2025} also leverages optical-flow priors for long-form video understanding, using them to refine frame hierarchy and prune redundant visual tokens. However, such approaches typically rely on an external flow estimator or extra motion backbone and require a separate preprocessing stage, making the pipeline heavier and less tightly integrated with downstream reasoning. In contrast, we use an estimator-free, direction-preserving motion source that integrates explicit temporal evidence directly into video MLLM encoding and generation, without a separate flow extraction stage.

Our method is also related to dual-branch architectures and feature-fusion designs. In traditional computer vision, dual-backbone, multi-branch, and cross-attention-based fusion paradigms have been widely explored~\cite{twostreamfusion2016,motionformer2021}. Building on this structural perspective, we construct a dual-branch visual encoder in which one ViT models static appearance and the other models explicit motion, and connect them through aligned inter-frame cross-attention before language reasoning. Unlike approaches that enhance temporal sensitivity within a single RGB stream, we preserve a dedicated motion pathway and inject its evidence into visual reasoning before the LLM.

Finally, our training design is related to reinforcement learning methods for LLMs and MLLMs. In particular, recent reasoning models suggest that a supervised cold-start stage followed by GRPO-style optimization can provide an effective training recipe~\cite{deepseekr1_2025,deepseekmath2024}. In video reasoning, however, vanilla GRPO remains limited because it does not by itself encourage sufficient temporal information use~\cite{videor1_2025}. Our work follows this broader training lineage, but targets temporal order sensitivity more explicitly than generic post-SFT RL.

\section{Method}
\label{sec:method}

COMET addresses video temporal reasoning through two complementary pillars: a temporally grounded architecture and a direction-aware optimization strategy. Architecturally, we convert implicit pixel change into explicit temporal evidence using Taylor differences, and let appearance and temporal signals interact across aligned frames before the LLM. For optimization, we train the model with a direction-aware objective matched to temporal order sensitivity. This section describes how these components form one coherent temporal modeling pipeline.

\subsection{Method Overview}
\label{subsec:method_overview}

As summarized in Fig.~\ref{fig:overview}, our framework treats temporal evidence as a first-class visual signal rather than leaving it to be recovered implicitly from sparse RGB snapshots. Our method is built around two linked pillars: an architecture that explicitly represents frame-to-frame change and injects it into appearance reasoning through inter-frame interaction, and a direction-aware training strategy that encourages the model to preserve temporal order sensitivity.

\subsection{Explicit Temporal Representation}
\label{subsec:taylor}

\begin{figure*}[t]
  \centering
  \includegraphics[width=0.88\textwidth]{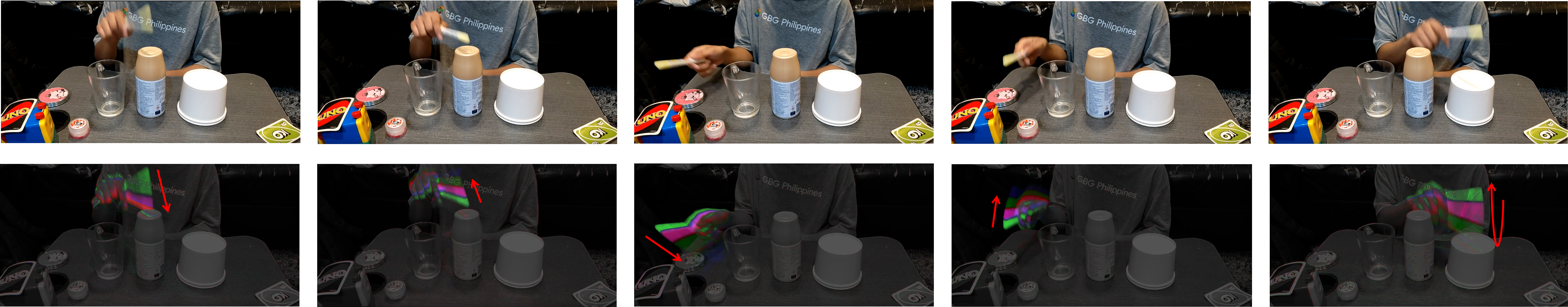}
  \caption{Taylor-style motion maps preserve temporal direction: reversing the video flips motion direction while largely preserving appearance.}
  \Description{A qualitative comparison of forward and reversed videos showing the directional information captured by Taylor-style motion maps.}
  \label{fig:taylor_case}
\end{figure*}

We begin from the representation level. Rather than relying on sparse RGB snapshots and expecting temporal information to be recovered implicitly, we explicitly encode frame-to-frame change before high-level reasoning. We do not use optical flow as an auxiliary modality, since it typically requires an external estimator~\cite{raft2020}, adds preprocessing cost, and introduces estimator-specific errors that are decoupled from the visual encoder. Instead, we use a Taylor-style representation built from grayscale frame differences, following prior work showing that Taylor videos provide a lightweight and effective motion-focused temporal representation~\cite{taylorvideo2024}. This choice aligns well with our architectural design: its main advantages are that it is lightweight, estimator-free, and preserves motion direction. Taylor features directly expose direction-preserving frame-to-frame change, rather than leaving motion to be inferred implicitly from sparse RGB snapshots, providing the motion branch with a lightweight yet explicit temporal signal that naturally supports both appearance-motion fusion and direction-aware training. Let $G_t$ denote the grayscale version of frame $I_t$. We recursively define temporal differences as
\begin{equation}
\mathbf{D}_t^{(0)} = G_t,\qquad
\mathbf{D}_t^{(k)} = \mathbf{D}_{t+1}^{(k-1)}-\mathbf{D}_t^{(k-1)},\quad k\ge 1.
\end{equation}
Abstracting the implementation over a temporal window of length $W$, we write the 3-channel Taylor response as
\begin{equation}
\mathbf{T}_t^{(c)}
=
\frac{1}{W}
\sum_{\delta=0}^{W-1}\sum_{b=0}^{2}
\frac{\left(G_{t+\delta}-G_t\right)^b}{b!}\,
\mathbf{D}_t^{(b+c)},
\qquad c\in\{1,2,3\},
\end{equation}
and use the Taylor video actually fed to ViT-B as a grayscale-blended map
\begin{equation}
\Delta_t = (1-\gamma)\,\mathrm{Norm}\!\left([\mathbf{T}_t^{(1)},\mathbf{T}_t^{(2)},\mathbf{T}_t^{(3)}]\right)
+ \gamma\, G_t\mathbf{1},
\qquad \gamma = 0.33,
\end{equation}
where $\mathrm{Norm}(\cdot)$ denotes per-channel clipping and rescaling, and $G_t\mathbf{1}$ replicates the grayscale frame across three channels. This design preserves coarse appearance context through the grayscale component while still making temporal change explicit through the Taylor response. Fig.~\ref{fig:taylor_case} shows this directional property qualitatively: reversing the video flips the Taylor motion pattern while largely preserving appearance. In our framework, the role of Taylor is to serve as a lightweight motion source that exposes a signed temporal signal while remaining easy to integrate into the video MLLM pipeline.
A key property of Taylor features is that they preserve motion directionality, which we exploit directly in TC-GRPO (Section~\ref{subsec:training}).

\subsection{Dual-Branch Visual Encoding}
\label{subsec:dual_branch}

With this representation in place, we next introduce an encoding design that keeps appearance and temporal evidence distinct while allowing them to interact across frames. We build a dual-branch visual encoder consisting of two ViTs with identical architecture but different inputs and roles:
\begin{equation}
\mathbf{F}^{A}_{l} = \text{ViT-A}(I;\theta_A)\vert_{l},\qquad
\mathbf{F}^{B}_{l} = \text{ViT-B}(\Delta;\theta_B)\vert_{l},
\end{equation}
where $\mathbf{F}^{A}_{l}$ and $\mathbf{F}^{B}_{l}$ denote the layer-$l$ features from the appearance and motion branches, respectively.

ViT-A encodes RGB frames and serves as the appearance branch, preserving the strong spatial prior of the original image-pretrained backbone. ViT-B is initialized from the same pretrained weights but processes Taylor inputs, making it the temporal branch. This separation is central to our design: RGB should answer ``what is present,'' while Taylor should answer ``what changes over time.'' The goal is not merely to add another branch, but to preserve explicit temporal evidence so it can later interact with appearance features at aligned frame positions.

We extract and fuse features at four DeepStack-aligned levels, $l\in\{5,11,17,24\}$, and denote the fused features by $\hat{\mathbf{F}}_l$. The fused features at levels 5, 11, and 17 are used as intermediate multi-level visual inputs to the LLM, while the deepest fused feature, $\hat{\mathbf{F}}_{24}$, serves as the final-level visual embedding. This design lets shallow fusion capture local motion patterns while deeper fusion carries higher-level temporal semantics into the LLM.

\begin{figure*}[t]
  \centering
  \includegraphics[width=0.82\textwidth]{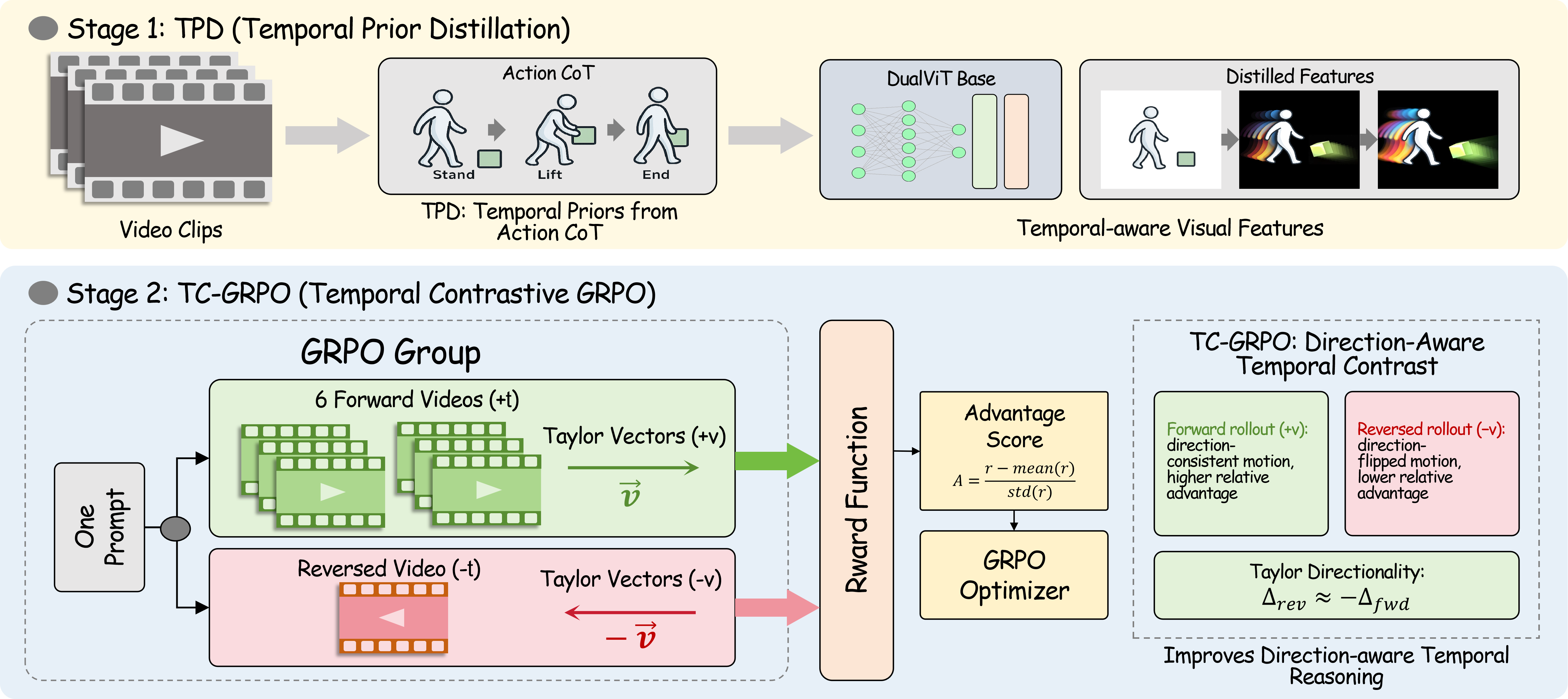}
  \caption{Temporal-specialized training pipeline with TPD and TC-GRPO.}
  \Description{An overview of the two-stage training pipeline, consisting of temporal prior distillation and temporal contrast GRPO.}
  \label{fig:training}
\end{figure*}

\subsection{Appearance-Motion Fusion}
\label{subsec:fusion}

The core interaction module is a cross-attention fusion block applied at each level. Its role is to create explicit inter-frame interaction rather than simply concatenate or average appearance and temporal features. For each spatial position $s$, we first project both branches into a bottleneck space:
\begin{equation}
\tilde{\mathbf{F}}^{A}_{l,s}=\mathrm{ProjDown}(\mathbf{F}^{A}_{l,s}),\qquad
\tilde{\mathbf{F}}^{B}_{l,s}=\mathrm{ProjDown}(\mathbf{F}^{B}_{l,s}).
\end{equation}
We then construct RGB queries and Taylor keys/values from normalized bottleneck features:
\begin{equation}
\mathbf{Q}_{s} = W_q\,\mathrm{LN}(\tilde{\mathbf{F}}^{A}_{l,s}),\qquad
\left[\mathbf{K}_{s}, \mathbf{V}_{s}\right] = W_{kv}\,\mathrm{LN}(\tilde{\mathbf{F}}^{B}_{l,s}).
\end{equation}
We then perform temporal cross-attention with an explicit temporal bias:
\begin{equation}
\mathbf{A}_{s}=
\mathrm{softmax}\!\left(
\frac{\mathbf{Q}_{s}\mathbf{K}_{s}^{\top}}{\sqrt{d_k}}+\mathbf{B}_{s,:}
\right),
\qquad
\mathbf{z}_{s}=\mathbf{A}_{s}\mathbf{V}_{s}.
\end{equation}
Finally, after reassembling $\mathbf{z}_{s}$ across all spatial positions into $\mathbf{z}_{l}$, the fused motion signal is projected back and injected into the RGB branch through a residual path:
\begin{equation}
\hat{\mathbf{F}}_l =
\mathbf{F}^{A}_{l} +
\tanh(\alpha_l)\cdot
\mathrm{RMSNorm}\!\left(\mathrm{ProjUp}(\mathbf{z}_{l})\right).
\end{equation}
Here, RGB features act as the main carrier signal and Taylor features provide motion evidence to be injected into them, so the model always retains a clean appearance anchor. After reshaping features into spatiotemporal tensors, we attend over time for each spatial position independently, so appearance tokens retrieve temporal evidence from the same location across frames rather than from unrelated tokens. This interaction pattern is important because it ties motion cues to aligned visual content instead of treating temporal information as a global residual. Cross-attention is performed in a bottleneck dimension rather than the full ViT hidden size, which reduces cost while preserving the effectiveness of appearance-motion interaction.

\paragraph{Temporal Attention Bias (TAB)}
Plain cross-attention still treats all frames at a location relatively uniformly. To bias fusion toward motion-salient moments, we add a temporal attention bias directly to the attention logits. Let $\bar{\mathbf{k}}_{s,t}$ denote the normalized Taylor key at spatial position $s$ and frame $t$, and let $\bar{\mathbf{k}}_{s,\cdot}=\frac{1}{T}\sum_t \bar{\mathbf{k}}_{s,t}$ be its temporal mean. The bias entry added to $\mathbf{B}_{s,:}$ is
\begin{equation}
B_{s,t}
=
\lambda\,
\underbrace{\|\bar{\mathbf{k}}_{s,t}\|_2}_{a_{s,t}}
\underbrace{\left(\frac{1}{D}\sum_d \mathrm{Var}_{t}(\bar{\mathbf{k}}_{s,t,d})\right)}_{v_s}
\underbrace{\|\bar{\mathbf{k}}_{s,t}-\bar{\mathbf{k}}_{s,\cdot}\|_2}_{d_{s,t}}.
\end{equation}
For brevity, we denote the product $a_{s,t}v_sd_{s,t}$ by $\rho_{s,t}$, so that $B_{s,t}=\lambda\rho_{s,t}$. Here, $a_{s,t}$ measures instantaneous motion strength, $v_s$ measures temporal activity at location $s$, and $d_{s,t}$ measures how distinctive frame $t$ is relative to the local temporal baseline $\bar{\mathbf{k}}_{s,\cdot}$. All three terms are read directly from Taylor features already computed in the fusion path, so TAB introduces no auxiliary scorer or extra visual branch. These three indicators capture complementary aspects of motion saliency, and their product $\rho_{s,t}$ assigns greater attention bias to Taylor responses that are simultaneously strong, temporally active, and distinctive from the local temporal baseline. TAB therefore steers fusion toward motion-relevant frame-location pairs instead of allowing temporally static evidence to dominate the aggregation.

To keep the fusion path stable during Stage\,I, we use a learnable residual scaling factor $\tanh(\alpha_l)$ with $\alpha_l$ initialized to zero, so the module starts from identity and motion injection increases only as training opens the gate. We further apply RMSNorm before residual addition to control the fusion magnitude. These are stability-oriented design choices rather than standalone innovations.

\subsection{Direction-Aware Temporal Optimization}
\label{subsec:training}

The final piece of the story is optimization. Architecture alone does not guarantee that the model will truly rely on temporal evidence: a model may still answer correctly through dataset bias or language priors. We therefore use a two-stage training paradigm that is explicitly matched to the structure of our temporal representation, as illustrated in Fig.~\ref{fig:training}.

\paragraph{Stage 1: Temporal Prior Distillation (TPD)}
As shown in Fig.~\ref{fig:training}, the first stage supervises the model with chain-of-thought video reasoning data so that temporal cues are distilled into the appearance-temporal encoder, establishing temporal priors for downstream language reasoning. Given an input pair $(x,\mathcal{V})$ and a target reasoning-answer sequence $y^\star$, we optimize the standard autoregressive objective
\begin{equation}
\mathcal{L}_{\text{TPD}}
=
-\sum_{n}\log p_{\theta}\!\left(y^\star_n \mid x,\mathcal{V}, y^\star_{<n}\right).
\end{equation}
We keep ViT-A frozen and update ViT-B, the fusion modules, and the LLM during this stage. This allows the temporal pathway to learn and inject Taylor features while the language model adapts to the newly introduced temporal semantics. This stage teaches the model to associate visual change with explicit temporal reasoning traces rather than only final answers, thereby injecting temporal priors before reinforcement learning.

\begin{figure*}[t]
  \centering
  \includegraphics[width=\textwidth]{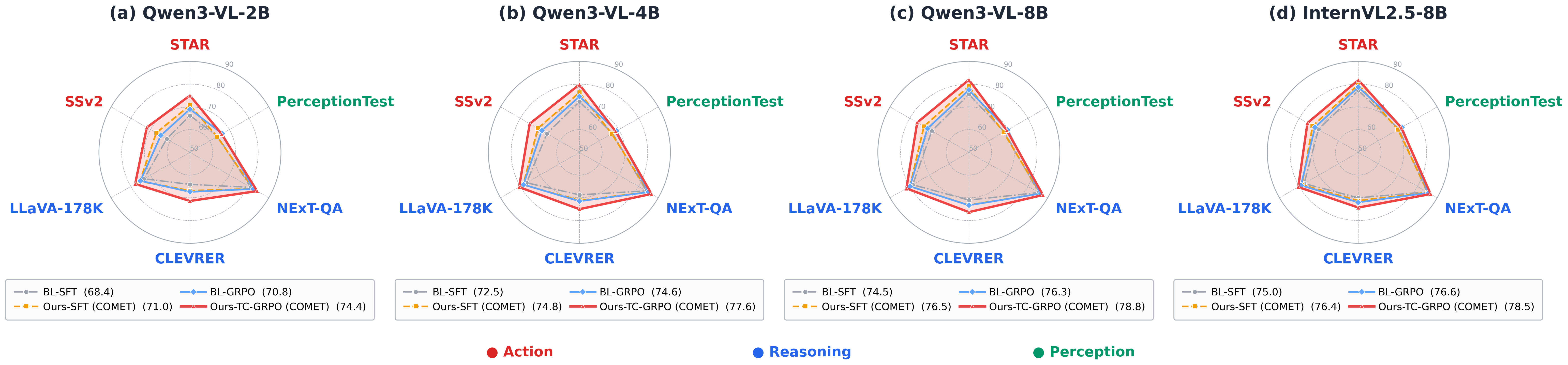}
  \caption{Radar plot of Table~\ref{tab:main}. Colors denote \textcolor{red}{Action}, \textcolor{blue}{Reasoning}, and \textcolor{teal}{Perception}.}
  \Description{A radar chart comparing model performance across action, reasoning, and perception task categories.}
  \label{fig:radar}
\end{figure*}

\paragraph{Stage 2: Temporal Contrast GRPO (TC-GRPO)}
As illustrated in Fig.~\ref{fig:training}, TPD teaches the model to produce temporal reasoning traces, but it does not force the model to truly rely on temporal direction. To address this gap, we exploit the directionality of Taylor features: Taylor frames leave colored trailing residuals resembling \ColorfulCOMET, reflecting motion direction and temporal order, and Fig.~\ref{fig:taylor_case} shows this directionality qualitatively under video reversal. In TC-GRPO, forward and reversed videos are placed in the same reinforcement-learning exploration space. The reversed condition is not merely a negative example; it forces the model to decide whether a video should be interpreted in its original or reversed order. To make that decision, the model is naturally driven to rely more on the directional cues carried by Taylor features. As a result, TC-GRPO strengthens temporal reasoning while also teaching the model to recognize directional Taylor frames and use motion-encoding information more effectively.

We optimize this process with GRPO's group-relative advantage. For each prompt $(x,\mathcal{V})$, we sample $N_f$ forward-video and $N_r$ reversed-video responses into a single group ($N_r \ll N_f$), keeping the textual question and reference answer fixed and forward samples dominant. This forward-dominant grouping preserves the main forward-video answering objective, while a small reversed subset is enough to make temporal direction matter during exploration:
\begin{equation}
\begin{aligned}
G^{\text{fwd}} &= \left[y_1^{\text{fwd}},\ldots,y_{N_f}^{\text{fwd}}\right], \quad
y_i^{\text{fwd}} \sim \pi_{\theta_{\mathrm{old}}}(\cdot\mid x,\mathcal{V}), \\
G^{\text{rev}} &= \left[y_1^{\text{rev}},\ldots,y_{N_r}^{\text{rev}}\right], \quad
y_j^{\text{rev}} \sim \pi_{\theta_{\mathrm{old}}}(\cdot\mid x,\mathcal{V}^{\text{rev}}), \\
G &= G^{\text{fwd}} \cup G^{\text{rev}}.
\end{aligned}
\end{equation}
We use answer correctness as the reward, with $R_i=1$ for a correct answer and $0$ otherwise. GRPO's group-relative advantage plays the role of an optimization interface here: it writes the reward differences formed during exploration into the standard policy update, leveraging direction-sensitive Taylor features so that the model automatically learns directional motion-encoding information during RL exploration and improves temporal capability. The TC-GRPO loss follows the clipped surrogate form:
\begin{equation}
\begin{aligned}
\mathcal{L}_{\text{TC-GRPO}}
&= -\frac{1}{|G|}
\sum_{i=1}^{|G|}
\left[
\frac{1}{|y_i|}\sum_{m=1}^{|y_i|}
\min\!\Big(
r_{i,m}\hat{A}_i,\,
\mathrm{clip}\!\left(r_{i,m},1-\epsilon_c,1+\epsilon_c\right)\hat{A}_i
\Big)\right], \\
\hat{A}_i &= \frac{R_i-\mu_G}{\sigma_G+\epsilon},\qquad
r_{i,m} =
\frac{\pi_{\theta}\!\left(y_{i,m}\mid x,\mathcal{V}_i,y_{i,<m}\right)}
{\pi_{\theta_{\mathrm{old}}}\!\left(y_{i,m}\mid x,\mathcal{V}_i,y_{i,<m}\right)}.
\end{aligned}
\end{equation}
where $\mathcal{V}_i\!=\!\mathcal{V}$ for forward responses and $\mathcal{V}_i\!=\!\mathcal{V}^{\text{rev}}$ for reversed ones. In Stage\,II, we freeze the visual pathway and update only the LLM. This still matches our temporal-first logic: Stage\,I has already learned a stable temporal prior in ViT-B and the fusion modules, so RL should optimize how the language model reads and uses that fixed temporal evidence, rather than letting high-variance reward updates distort the motion encoder itself.

\begin{table*}[t]
\centering
\caption{Main results on six benchmarks and subsets. Stage\,0: pretrained; I: SFT/TPD; II: GRPO/TC-GRPO; X: external baselines. $^\dagger$Reproduced under our setup. Best in \textbf{bold}, second-best \underline{underlined}. \textcolor{red}{Action}=avg(STAR, SSv2), \textcolor{blue}{Reasoning}=avg(NExT-QA, CLEVRER, LLaVA-178K), \textcolor{teal}{Perception}=PercepTest.}
\label{tab:main}
\small
\setlength{\tabcolsep}{2.6pt}
\begin{tabular}{ll@{\hspace{7pt}}|@{\hspace{7pt}}c@{\hspace{7pt}}|@{\hspace{7pt}}cccccc@{\hspace{8pt}}|@{\hspace{6pt}}cccc}
\toprule
& Method & Stage & \textcolor{red}{STAR} & \textcolor{teal}{PercepTest} & \textcolor{blue}{NExT-QA} & \textcolor{blue}{CLEVRER} & \textcolor{blue}{LLaVA-178K} & \textcolor{red}{SSv2} & \textcolor{red}{Action} & \textcolor{blue}{Reasoning} & \textcolor{teal}{Perception} & Avg \\
\midrule
\multicolumn{13}{l}{\textit{Qwen3-VL-2B}} \\
\midrule
& Pretrained & 0 & 60.7 & 60.3 & 78.8 & 59.9 & 69.8 & 55.8 & 58.3 & 69.5 & 60.3 & 64.2 \\
& BL-SFT & I & 66.2 & 64.1 & 80.8 & 64.1 & 73.3 & 61.8 & 64.0 & 72.7 & 64.1 & 68.4 \\
& Ours-TPD & I & 70.7 & 63.8 & 82.1 & 66.9 & 75.4 & \underline{67.1} & \underline{68.9} & 74.8 & 63.8 & 71.0 \\
& BL-GRPO & II & 69.0 & 66.4 & 82.2 & 67.4 & 75.1 & 64.8 & 66.9 & 74.9 & 66.4 & 70.8 \\
& Flow4Agent (SAMFlow)$^\dagger$~\cite{flow4agent2025} & X & \underline{70.9} & \textbf{66.8} & \underline{82.7} & \underline{68.7} & \underline{75.9} & 66.9 & \underline{68.9} & \underline{75.8} & \textbf{66.8} & \underline{72.0} \\
& TempFlex$^\dagger$~\cite{tempflex2025} & X & 70.6 & \underline{66.7} & 82.5 & 68.4 & 75.7 & 66.8 & 68.7 & 75.5 & \underline{66.7} & 71.8 \\
& \textbf{Ours-TC-GRPO} & II & \textbf{75.0} & 66.1 & \textbf{84.1} & \textbf{71.4} & \textbf{77.8} & \textbf{71.9} & \textbf{73.5} & \textbf{77.8} & 66.1 & \textbf{74.4} \\
\midrule
\multicolumn{13}{l}{\textit{Qwen3-VL-4B}} \\
\midrule
& Pretrained & 0 & 67.5 & 63.5 & 82.1 & 65.1 & 73.8 & 61.3 & 64.4 & 73.7 & 63.5 & 68.9 \\
& BL-SFT & I & 72.3 & 66.8 & 83.9 & 68.7 & 76.8 & 66.5 & 69.4 & 76.5 & 66.8 & 72.5 \\
& Ours-TPD & I & \underline{76.3} & 66.4 & 85.0 & 71.2 & 78.5 & 71.1 & 73.7 & 78.2 & 66.4 & 74.8 \\
& BL-GRPO & II & 74.6 & 68.8 & 85.0 & 71.5 & 78.4 & 69.1 & 71.9 & 78.3 & 68.8 & 74.6 \\
& Flow4Agent (SAMFlow)$^\dagger$~\cite{flow4agent2025} & X & \underline{76.3} & \textbf{69.1} & \underline{85.4} & \underline{72.5} & \underline{79.0} & \underline{71.3} & \underline{73.8} & \underline{79.0} & \textbf{69.1} & \underline{75.6} \\
& TempFlex$^\dagger$~\cite{tempflex2025} & X & 75.9 & \underline{69.0} & 85.2 & 72.3 & 78.8 & 71.2 & 73.6 & 78.8 & \underline{69.0} & 75.4 \\
& \textbf{Ours-TC-GRPO} & II & \textbf{79.8} & 68.4 & \textbf{86.6} & \textbf{75.0} & \textbf{80.6} & \textbf{75.2} & \textbf{77.5} & \textbf{80.7} & 68.4 & \textbf{77.6} \\
\midrule
\multicolumn{13}{l}{\textit{Qwen3-VL-8B}} \\
\midrule
& Pretrained & 0 & 71.3 & 65.2 & 83.9 & 68.0 & 76.0 & 64.3 & 67.8 & 76.0 & 65.2 & 71.5 \\
& BL-SFT & I & 75.5 & 68.0 & 85.4 & 71.0 & 78.5 & 68.8 & 72.2 & 78.3 & 68.0 & 74.5 \\
& Ours-TPD & I & \underline{79.0} & 67.6 & 86.3 & 73.2 & 79.9 & 72.8 & \underline{75.9} & 79.8 & 67.6 & 76.5 \\
& BL-GRPO & II & 77.4 & 69.7 & 86.3 & 73.3 & 79.9 & 71.0 & 74.2 & 79.8 & 69.7 & 76.3 \\
& Flow4Agent (SAMFlow)$^\dagger$~\cite{flow4agent2025} & X & 78.7 & \textbf{70.0} & \underline{86.6} & \underline{74.1} & \underline{80.3} & \underline{73.1} & \underline{75.9} & \underline{80.3} & \textbf{70.0} & \underline{77.1} \\
& TempFlex$^\dagger$~\cite{tempflex2025} & X & 78.4 & \underline{69.9} & 86.4 & 73.8 & 80.1 & 73.0 & 75.7 & 80.1 & \underline{69.9} & 76.9 \\
& \textbf{Ours-TC-GRPO} & II & \textbf{81.9} & 69.3 & \textbf{87.6} & \textbf{76.3} & \textbf{81.7} & \textbf{76.2} & \textbf{79.1} & \textbf{81.9} & 69.3 & \textbf{78.8} \\
\midrule
\multicolumn{13}{l}{\textit{InternVL2.5-8B}} \\
\midrule
& Pretrained & 0 & 73.7 & 68.3 & 83.4 & 67.6 & 75.8 & 66.8 & 70.2 & 75.6 & 68.3 & 72.6 \\
& BL-SFT & I & 77.2 & 70.6 & 84.8 & 70.0 & 77.5 & 70.2 & 73.7 & 77.4 & 70.6 & 75.0 \\
& Ours-TPD & I & \underline{79.8} & 70.1 & 85.3 & 71.3 & 78.6 & \underline{73.2} & \underline{76.5} & 78.4 & 70.1 & 76.4 \\
& BL-GRPO & II & 78.7 & \textbf{72.1} & \underline{85.6} & \underline{72.0} & \underline{79.1} & 72.0 & 75.3 & \underline{78.9} & \textbf{72.1} & \underline{76.6} \\
& \textbf{Ours-TC-GRPO} & II & \textbf{81.8} & \underline{71.8} & \textbf{86.7} & \textbf{74.3} & \textbf{80.5} & \textbf{75.8} & \textbf{78.8} & \textbf{80.5} & \underline{71.8} & \textbf{78.5} \\
\bottomrule
\end{tabular}
\end{table*}

\section{Experiments}
\label{sec:experiments}

\subsection{Experimental Setup}
\label{subsec:setup}

We apply COMET to two representative video MLLM families: Qwen3-VL~\cite{qwen3vl2025} (2B/4B/8B), which uses DeepStack multi-level injection, and InternVL2.5-8B~\cite{internvl25_2024}, which follows a standard ViT-MLP-LLM architecture. ViT-B is initialized from each model's pretrained ViT-A weights.

We use Video-R1~\cite{videor1_2025} throughout. Stage\,I (TPD) trains on the chain-of-thought reasoning split (${\sim}$120K samples); Stage\,II (TC-GRPO) uses the RL split (${\sim}$40K samples filtered for temporal sensitivity). Frames are sampled at stride\,6 in both stages (frame count varies with video length). Stage\,I keeps ViT-A frozen while updating ViT-B, the fusion modules, and the LLM. Stage\,II freezes the entire visual pathway and updates only the LLM, using a group size of 7 ($N_f\!=\!6$ forward, $N_r\!=\!1$ reversed) and a clip ratio $\epsilon_c\!=\!0.2$. All experiments are conducted on 8$\times$A100 (80\,GB).

We evaluate on six benchmarks and evaluation subsets spanning three task categories: \textcolor{red}{\textbf{Action}} --- STAR~\cite{star2021} (situated reasoning) and SSv2~\cite{something2017} (compositional temporal recognition); \textcolor{blue}{\textbf{Reasoning}} --- NExT-QA~\cite{nextqa2021} (causal and temporal QA), CLEVRER~\cite{clevrer2020} (compositional video reasoning), and LLaVA-178K~\cite{llavavideo2025} (an open-ended QA subset derived from the data release accompanying LLaVA-Video); \textcolor{teal}{\textbf{Perception}} --- PerceptionTest~\cite{perceptiontest2023} (general video perception). We randomly sample 1,000 examples per benchmark or subset and report accuracy under the same evaluation protocol for all methods. For SSv2, we add a 5-way multiple-choice prompt per sample (1 ground-truth template + 4 random distractors), and use our checker to extract the predicted option letter from the final \texttt{<answer>} tag, with a letter-based fallback, before matching it against the ground-truth choice.

We compare against the zero-shot backbone, standard SFT and GRPO baselines, and two representative temporal baselines, Flow4Agent (SAMFlow)~\cite{flow4agent2025} and TempFlex~\cite{tempflex2025}. For fair comparison on Qwen3-VL, we re-implement Flow4Agent with SAMFlow motion priors and adapt TempFlex to the same backbone under the same training and evaluation pipeline.

\subsection{Overall Results and Temporal Sensitivity}
\label{subsec:overall_results}

Table~\ref{tab:main} summarizes the overall results, and Fig.~\ref{fig:radar} provides a category-level view of the same trend. Each training stage contributes a distinct improvement. Standard video-QA fine-tuning (BL-SFT) improves the pretrained models by 2.4--4.2\,pp. Applying COMET with TPD provides a further 1.4--2.6\,pp by making temporal evidence explicit, and TC-GRPO yields cumulative gains of 5.9--10.2\,pp over the pretrained models across scales and backbones.

Against strong temporal baselines, Ours-TC-GRPO consistently performs best across all three Qwen3-VL scales. We compare against Flow4Agent (SAMFlow)~\cite{flow4agent2025}, an optical-flow-prior baseline, and TempFlex~\cite{tempflex2025}, a temporal-fusion baseline, under the same SFT + standard GRPO pipeline. The gains over these baselines are moderate but consistent, while COMET remains estimator-free and uses a single-stage visual pipeline rather than a separate motion-estimation stage. The same temporal advantage also transfers to InternVL2.5-8B, which follows a standard ViT-MLP-LLM design rather than Qwen3-VL's DeepStack pathway. Without multi-level DeepStack fusion, COMET reduces to a single main-level fusion module, yet still achieves the best average accuracy, indicating that the core temporal signal transfers across architectures.

\paragraph{Category-level pattern}
As shown in the rightmost columns of Table~\ref{tab:main} and the radar visualization in Fig.~\ref{fig:radar}, COMET's advantage follows a consistent \textbf{\textcolor{red}{Action} $>$ \textcolor{blue}{Reasoning} $>$ \textcolor{teal}{Perception}} hierarchy across all scales. At 8B, Action improves by +4.9\,pp and Reasoning by +2.1\,pp over BL-GRPO, while Perception shows a slight deficit ($-$0.4\,pp). This tradeoff is distinctive to COMET under our setup: Flow4Agent (SAMFlow) and TempFlex both improve on PerceptionTest slightly, whereas COMET concentrates its gains on Action and Reasoning, with only a small change on PerceptionTest. This pattern is consistent with our design goal of prioritizing explicit temporal change and directional cues rather than seeking uniform gains on all video tasks.

\subsection{Component Ablation}
\label{subsec:ablation}

We ablate COMET's core modules to quantify their cumulative contribution. Table~\ref{tab:component} shows the effect of first adding the temporal branch with appearance-motion fusion and then adding TAB on the 8B backbone at Stage\,I.

\begin{table}[h]
\centering
\caption{Component ablation of COMET (Qwen3-VL-8B, Stage\,I). $\Delta$ denotes the gain over the previous row.}
\label{tab:component}
\small
\setlength{\tabcolsep}{3pt}
\begin{tabular}{lcccccc|cc}
\toprule
Configuration & STAR & PT & NQA & CLV & LLaVA & SSv2 & Avg & $\Delta$ \\
\midrule
Baseline (BL-SFT) & 75.5 & \textbf{68.0} & 85.4 & 71.0 & 78.5 & 68.8 & 74.5 & --- \\
COMET w/o TAB & 78.4 & 67.7 & 86.1 & 72.8 & 79.6 & 72.1 & 76.1 & +1.6 \\
\textbf{Full COMET} & \textbf{79.0} & 67.6 & \textbf{86.3} & \textbf{73.2} & \textbf{79.9} & \textbf{72.8} & \textbf{76.5} & +0.4 \\
\bottomrule
\end{tabular}
\end{table}

The ablation mainly serves as a sanity check that each component is helpful. Most of the benefit comes from introducing the temporal branch together with appearance-motion fusion, while TAB provides a modest additional refinement in this setting. We therefore do not claim a large standalone gain for any single component; rather, the value of COMET is that these pieces combine into an estimator-free temporal pipeline that remains relatively lightweight.

\begin{figure}[t]
  \centering
  \includegraphics[width=\columnwidth]{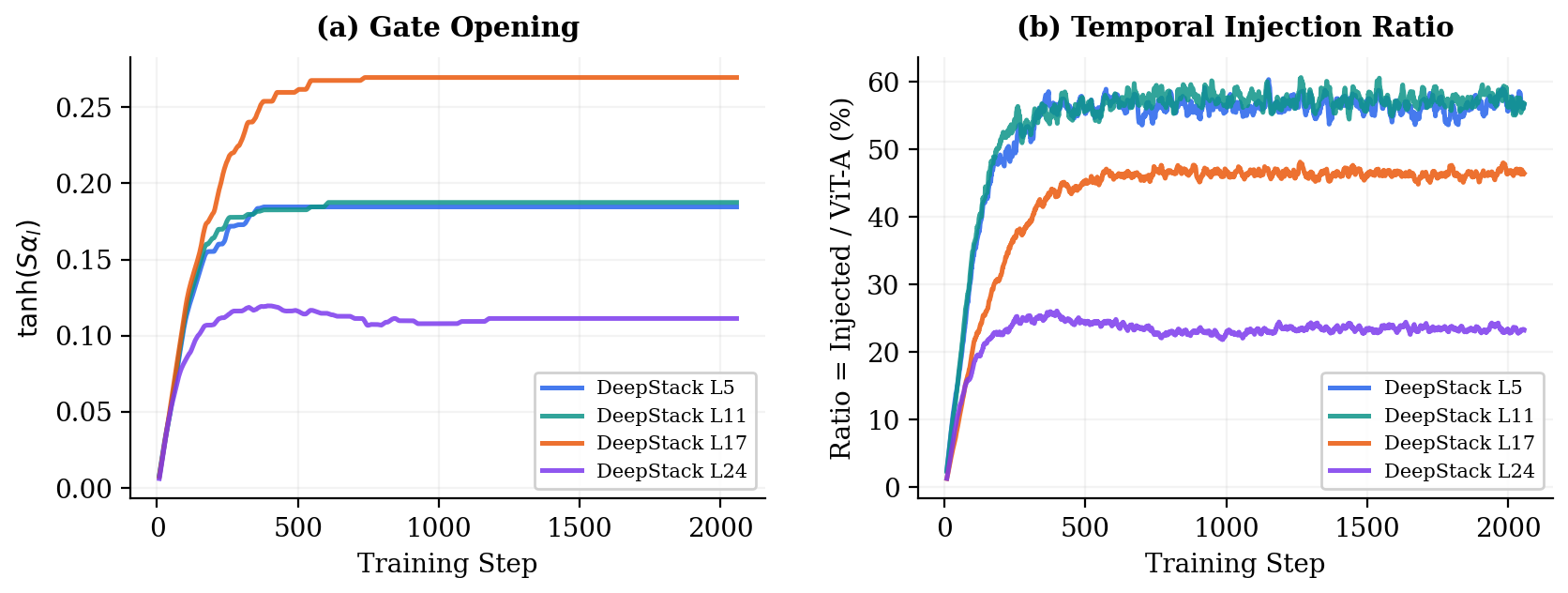}
  \caption{Fusion gate dynamics during Stage\,I. (a)~Gate opening $\tanh(\alpha_l)$ by level. (b)~Injection ratio relative to ViT-A features.}
  \Description{Two plots showing the evolution of fusion gates and motion-injection ratios during Stage I training.}
  \label{fig:gate_opening}
\end{figure}

\paragraph{Gate activation analysis}
Fig.~\ref{fig:gate_opening} visualizes the fusion gate dynamics during Stage\,I training. Consistent with $\alpha_l$ being initialized to zero in Eq.~(8), the fusion path starts from identity and the gates open gradually during TPD as the model learns to use temporal evidence, with deeper layers (L17) opening furthest. However, deeper layers also have larger ViT-A feature norms, which moderates their effective injection ratio and prevents any single level from dominating the temporal signal.

\begin{table}[h]
\vspace{10pt}
\centering
\caption{Parameter overhead on 8B-scale backbones. \textbf{Ours}: Taylor ViT-B core attention parameters and fusion modules. \textbf{Flow}: external optical-flow estimators.}
\label{tab:overhead}
\small
\setlength{\tabcolsep}{5pt}
\begin{tabular}{@{}l@{\quad}l|cc@{}}
\toprule
& Components & Qwen3-VL-8B & InternVL2.5-8B \\
\midrule
\multirow{2}{*}{\textbf{Ours}}
  & Taylor ViT-B & +143.3\,M\;(+1.6\,\%) & +100.7\,M\;(+1.2\,\%) \\
  & Fusion modules & +37.8\,M $\times$ 4\;(+1.7\,\%) & +28.3\,M $\times$ 1\;(+0.4\,\%) \\
\midrule
\multirow{2}{*}{\textbf{Flow}}
  & SAMFlow~\cite{samflow2024} & +650\,M\;(+7.4\,\%) & +650\,M\;(+8.0\,\%) \\
  & MegaFlow~\cite{megaflow2025} & +936\,M\;(+10.7\,\%) & +936\,M\;(+11.6\,\%) \\
\bottomrule
\end{tabular}
\vspace{-5pt}
\end{table}

\paragraph{Computational overhead}
Table~\ref{tab:overhead} reports the parameter overhead on two 8B-scale backbones. To avoid over-counting backbone-specific visual details in COMET, we use a conservative accounting for ViT-B and report its core attention parameters, while listing the fusion modules separately. The table therefore makes explicit how COMET's added cost is distributed between the Taylor branch and the fusion modules, while the flow-based baselines are represented by the external optical-flow estimators used to provide motion priors. This accounting keeps the comparison focused on the core temporal machinery without over-claiming backbone-dependent details.

\subsection{Direction-Aware Optimization Analysis}
\label{subsec:optimization_analysis}

To isolate the contribution of the optimization strategy from the architecture, we compare standard GRPO and TC-GRPO applied to the same full COMET architecture, starting from the identical Stage\,I checkpoint (Qwen3-VL-8B). Table~\ref{tab:grpo_ablation} presents the results.

\begin{table}[h]
\centering
\caption{Optimization ablation on COMET (Qwen3-VL-8B; group size\,=\,7). All variants start from the same Stage\,I checkpoint. $\Delta$ denotes the gain over Ours-TPD.}
\label{tab:grpo_ablation}
\small
\setlength{\tabcolsep}{3pt}
\begin{tabular}{lcccccc|cc}
\toprule
Optimization & STAR & PT & NQA & CLV & LLaVA & SSv2 & Avg & $\Delta$ \\
\midrule
Ours-TPD (Stage\,I) & 79.0 & 67.6 & 86.3 & 73.2 & 79.9 & 72.8 & 76.5 & --- \\
Std.\ GRPO (7+0) & 80.6 & 69.0 & 87.0 & 75.5 & 81.2 & 74.7 & 78.0 & +1.5 \\
\textbf{TC-GRPO (6+1)} & \textbf{81.9} & \textbf{69.3} & \textbf{87.6} & \textbf{76.3} & \textbf{81.7} & \textbf{76.2} & \textbf{78.8} & \textbf{+2.3} \\
TC-GRPO (5+2) & 81.3 & 69.1 & 87.3 & 75.9 & 81.5 & 75.3 & 78.4 & +1.9 \\
TC-GRPO (4+3) & 80.5 & 68.5 & 86.7 & 75.0 & 80.8 & 74.1 & 77.6 & +1.1 \\
\bottomrule
\end{tabular}
\vspace{15pt}
\end{table}

All GRPO variants improve over Stage\,I, but 6+1 performs best, reaching 78.8 average accuracy versus 78.0 for standard GRPO (7+0), 78.4 for 5+2, and 77.6 for 4+3. This suggests that a small number of reversed samples is already sufficient to make temporal direction matter during exploration, while increasing the reversed proportion places too much weight on the auxiliary reversed condition and weakens the main forward-video answering objective.

\begin{figure}[t]
  \centering
  \includegraphics[width=\columnwidth]{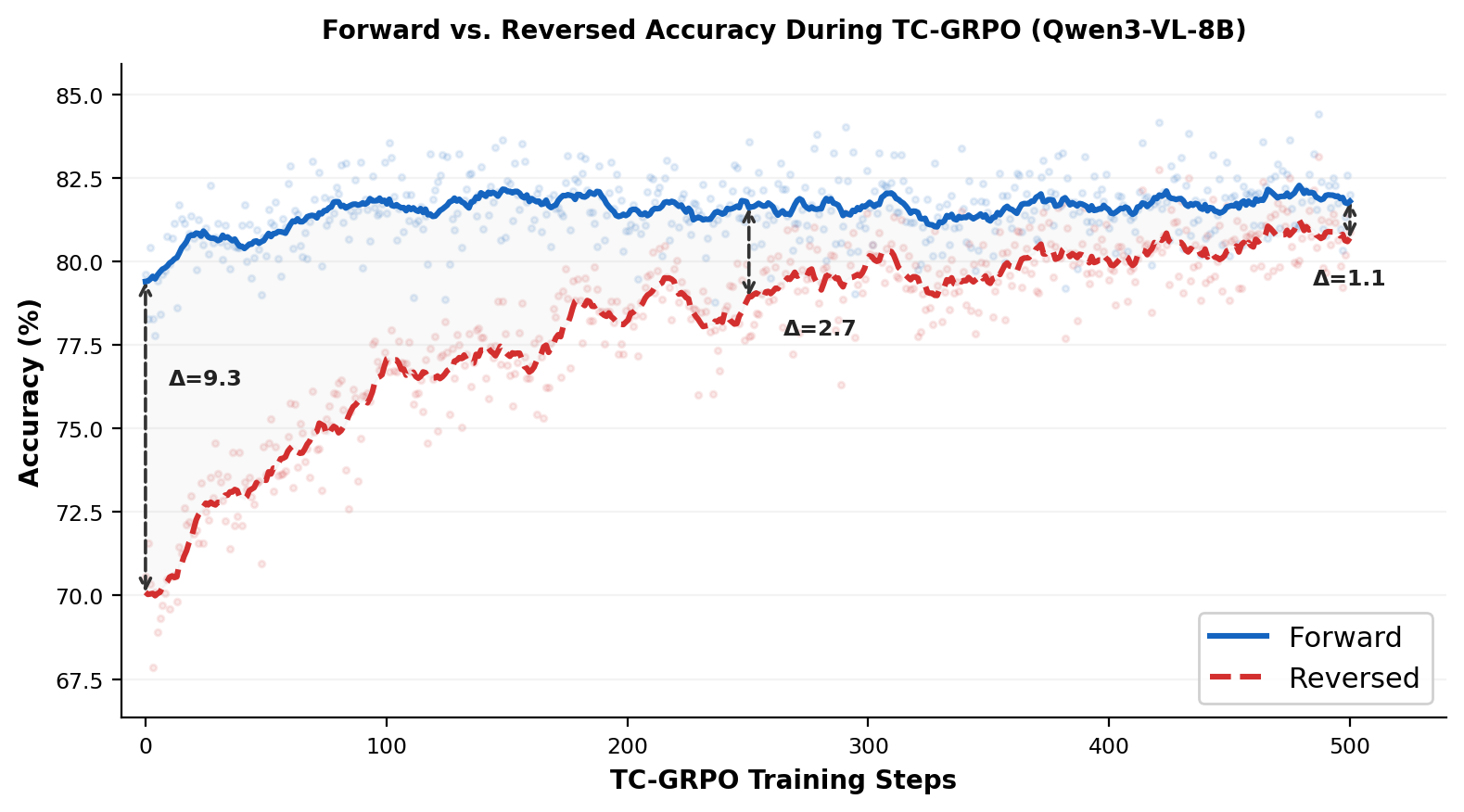}
  \caption{Forward vs.\ reversed accuracy during TC-GRPO training (Qwen3-VL-8B).}
  \Description{A line chart showing forward-video and reversed-video accuracy during TC-GRPO training.}
  \label{fig:fwdrev}
\end{figure}

Fig.~\ref{fig:fwdrev} further tracks forward and reversed video accuracy during TC-GRPO training. At step\,0, the clear forward--reversed gap shows that COMET-TPD already distinguishes temporal direction through Taylor features. During TC-GRPO, reversed-video accuracy rises steadily while forward-video accuracy remains strong, meaning that the model is not losing direction sensitivity; instead, it is learning how to handle the added reversed-order condition within the same exploration space. Together with the gain over standard GRPO in Table~\ref{tab:grpo_ablation}, this supports improved temporal robustness and direction-aware reasoning.

\section{Conclusion}
\label{sec:conclusion}

We present a temporally grounded video MLLM framework that addresses the underuse of temporal evidence in current pipelines. By making frame-to-frame change explicit, enabling inter-frame appearance-motion interaction, and directly optimizing temporal directionality, COMET improves video reasoning beyond sparse frame sampling and appearance-dominated shortcuts. Experiments on Qwen3-VL (three scales) and InternVL2.5 demonstrate that the core temporal signal---Taylor features and direction-aware optimization---transfers across model families, from DeepStack-based architectures to standard ViT-MLP-LLM designs. Future work will explore deeper fusion of the two ViT branches---e.g., weight sharing or progressive merging---to reduce the parameter footprint while enabling even tighter inter-frame interaction, and further validate this design on broader benchmarks and longer-form video reasoning settings.

\begin{acks}
This work was supported by the National Key Research and Development Program of China (Grant No. 2024YFE0203100), the Guangdong Provincial Key Laboratory of Ultra High Definition Immersive Media Technology (Grant No. 2024B1212010006), and the Shenzhen Science and Technology Program (Grant No. JCYJ20230807120800001).
\end{acks}

\bibliographystyle{ACM-Reference-Format}
\bibliography{references}

\end{document}